\pdfoutput=1
\documentclass[11pt]{article}

\usepackage[acceptedWithA]{tacl2021v1}
\usepackage{times}
\usepackage{latexsym}
\usepackage{todonotes}
\usepackage[T1]{fontenc}
\usepackage{booktabs, multirow} % for borders and merged ranges
\usepackage{soul}% for underlines
\usepackage[utf8]{inputenc}
\usepackage{float}
\usepackage{microtype}
\usepackage{caption}
\usepackage{subcaption}
\usepackage[inline]{enumitem}
\usepackage{inconsolata}
\usepackage{amsmath}
\usepackage[utf8]{inputenc}
\usepackage{devanagari}
\usepackage{lscape}
\usepackage[OT2, T1]{fontenc}
\usepackage[russian, english]{babel}
\usepackage{soul}

\title{Comparing Character- and Subword-level Pretrained Models\\ for Neural Machine Translation}

\title{Are Character-level Translations Worth the Wait? \\
Comparing Pretrained Character- and Subword-level \\ Models for Machine Translation}

\title{Are Character-level Translations Worth the Wait? \\
Comparing ByT5 and mT5 for Machine Translation}

\author{Lukas Edman \qquad  Gabriele Sarti \qquad Antonio Toral \\
\bf{Gertjan van Noord \qquad Arianna Bisazza}
\vspace{.2cm}
 \\ Center for Language and Cognition \\ 
 University of Groningen \vspace{.1cm}
 \\ {\tt \small\{j.l.edman, g.sarti, a.toral.ruiz, g.j.m.van.noord, a.bisazza\}@rug.nl}
}
\begin{document}
\maketitle
\begin{abstract}
Pretrained character-level and byte-level language models have been shown to be competitive with popular subword models across a range of Natural Language Processing (NLP) tasks.
However, there has been little research on their effectiveness for neural machine translation (NMT), particularly within the popular pretrain-then-finetune paradigm.
This work performs an extensive comparison across multiple languages and experimental conditions of character- and subword-level pretrained models (ByT5 and mT5, respectively) on NMT. 
We show the effectiveness of character-level modeling in translation, particularly in cases where fine-tuning data is limited.
In our analysis, we show how character models' gains in translation quality are reflected in better translations of orthographically similar words and rare words. While evaluating the importance of source texts in driving model predictions, we highlight word-level patterns within ByT5, suggesting an ability to modulate word-level and character-level information during generation. 
We conclude by assessing the efficiency tradeoff of byte models, suggesting their usage in non-time-critical scenarios to boost translation quality.
\end{abstract}

\section{Introduction}
Character-level and byte-level models\footnote{Byte models are often referred to under the broader category of character models in literature, however there are subtle differences. In this work, we exeriment only with byte-level models. We distinguish the two when we consider the differences to be notable, but otherwise continue to refer to byte models as character models.} have been a source of interest in Natural Language Processing (NLP) for many years, with the promise of tokenization-free systems able to process and generate text on a finer granularity. However, these systems have failed to become the dominant paradigm over subword-level models, despite comparable performances. This is likely because of the additional time and compute resources required, due to the longer input sequences used in character-based approaches. There are cases where character models have been shown to outperform subword models. However, these instances may be seen as niche (e.g. tasks that specifically require character information) or unrealistic (e.g. using data corrupted with synthetic noise~\citep{xue-etal-2022-byt5}). 

Additionally, previous studies presented only limited evaluations of these systems for popular tasks where character-level information could lead to major performance benefits. 
In this work, we conduct a comprehensive comparison of character and subword pretrained models on machine translation (MT). 
Despite the popularity of NMT, pretrained character models have not yet been thoroughly assessed for this task, with recent research on character models for MT focusing on models trained from scratch \cite{libovicky-etal-2022-dont, edman-etal-2022-subword}. We posit that these approaches can only reliably assess translation performance for high-resource languages, where the beneficial effects of multilingual pretraining were found to be less impactful \cite{liu-etal-2020-multilingual-denoising}. 

Character models can leverage more fine-grained information, which could be helpful in many challenging NMT settings, such as low-resource translation.
% Given the diversity of NMT settings, be it low or high-resource, similar or distant language pairs, or various writing scripts, the finer granularity of character models has the potential to benefit in many scenarios.
To validate this, we fine-tune the character model ByT5 \cite{xue-etal-2022-byt5} and its subword counterpart mT5 \cite{xue-etal-2021-mt5} for translation, for a variety of languages and settings. Among our findings, those standing out are: %ByT5 outperforms mT5 in many circumstances for translation, performing particularly well in low-resource situations.

\begin{figure}
    \centering
    \includegraphics[width=220pt]{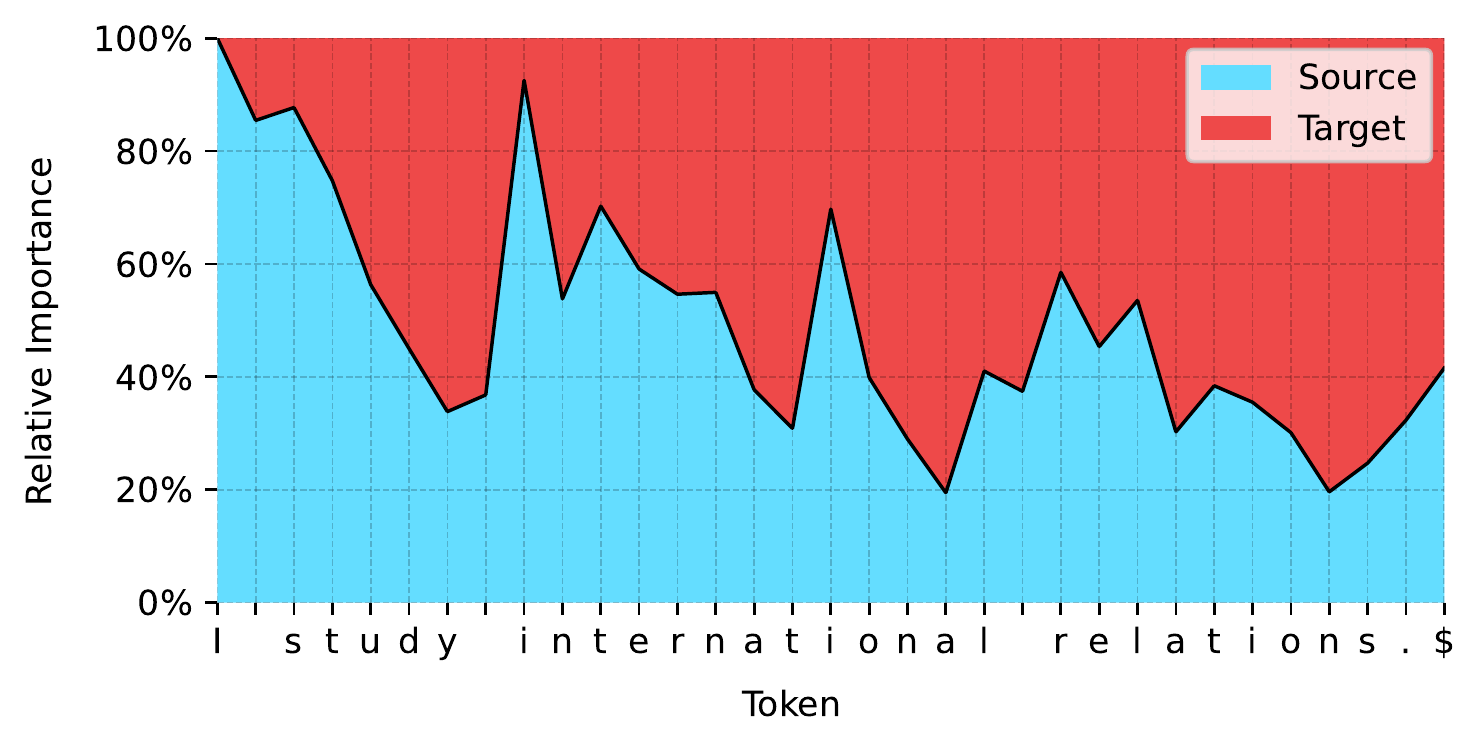}
    \caption{ByT5's overall source vs. target contribution for the German$\rightarrow$English translation: ``I study international relations.'' \$: end-of-sentence token. Peaks in source importance at the beginning of each word suggest word-level modeling.}
    \label{fig:attr_example}
    \vspace{-12pt}
\end{figure}

\begin{enumerate}[label=\textbf{(\arabic*)}, itemsep=-3pt]
    \item ByT5 generates higher quality translations than mT5 in general, especially when resources are limited.
    \item ByT5 shows better cross-lingual generalization than mT5 on average, and especially for high resource languages or low resource languages that are related to high resource ones.  
    \item Fine-tuning with many examples causes ByT5's translations in the zero-shot setting to degrade faster than mT5.
    \item ByT5 shows a cohesive word-level source importance pattern, suggesting a capacity to capture relationships beyond the character level (as shown in Figure \ref{fig:attr_example}). 
    \item When ByT5 outperforms mT5, ByT5 is also better at translating orthographically similar words and rare words.
\end{enumerate}

Our findings support the idea that in several realistic circumstances, and particularly for low-resource scenarios, character models are superior in terms of translation quality over their widely-used subword counterparts. In our analysis, we further show how these gains in quality connect to specific word properties, such as orthographic similarity and frequency.

\section{Related Work} \label{sect:rw}
Character-level models have long been of interest for use in machine translation, dating back to when statistical models were the dominant paradigm \cite{tiedemann-nakov-2013-analyzing}. At that time, character models were already competitive with word-level models, especially when training data was limited to $<\,$100k sentence pairs. \citet{durrani-etal-2010-hindi} also showed that character models were particularly adept at translating closely related languages such as Hindi and Urdu.
We note that, at the time, subword tokenizers such as BPE \cite{sennrich-etal-2016-neural} were not yet commonly used.

Neural approaches to character-level MT
%Character-level NMT models have also been researched, 
were first based on RNNs \cite{costa-jussa-fonollosa-2016-character, lee-etal-2017-fully}, with extensive work being done to compare character models to subword models (now equipped with BPE). For NMT, it was shown that character models using an RNN architecture (with and without a CNN for processing characters) perform equally to or better than subword models 
\cite{larriba2017traduccion,lee-etal-2017-fully,sennrich-2017-grammatical,cherry-etal-2018-revisiting}.  Their better translation performance could be attributed to their ability to handle complex morphology, rare or unseen words, and noisy input
\cite{jozefowicz2016exploring,kim2016character,belinkov2017synthetic,lee-etal-2017-fully,singh2017handling,durrani-etal-2019-one}. However the inefficiency of character models was already apparent, as they were slower than subword models by a considerable margin \cite{chung-etal-2016-character}.
Work had also been done in comparing byte and character level models, with little differences found, however these were mostly experimenting with Latin-scripted languages \cite{costa-jussa-etal-2017-byte}.

More recent work has looked at Transformers on the character and byte-level. \citet{xue-etal-2022-byt5} created ByT5 and compared it to its subword counterpart, mT5 \cite{xue-etal-2021-mt5}, which are also the models we focus on in this work. Their comparisons were however focused on either multilingual classification tasks or English-based generative tasks, but no multilingual generative tasks or machine translation.

Previous work analyzed character-level Transformers trained from scratch for NMT, not using the T5-based models. 
\citet{libovicky-etal-2022-dont} looked at ``vanilla'' character models (those without any compression of the sequence length prior to the computation in the Transformer), as well as \citet{lee-etal-2017-fully}'s, \citet{tay2021charformer}'s, and \citet{clark-etal-2022-canine}'s methods for compressing sequence length.
They conclude that these character-level models do not provide any benefits over subword models while being less efficient. 
There are two factors to note with these conclusions. Firstly, their experiments show similar performance, but they only experiment on high-resourced languages. Given prior work in RNNs mentioned above, this does not appear to be the application that would benefit most from character-level models.
Secondly, they introduce a two-step decoder to achieve character-level decoding from subword-level hidden states, however as \citet{edman-etal-2022-subword} points out, this decoder does not scale well to higher-resourced scenarios. This two-step decoder adds an additional layer of complexity to evaluating such models, as an ablation of the model's granularity and the model's decoding process would be necessary to fully understand the performance of each model. Added to the fact that there are no pretrained models using this sequence length compression that are comparable to mT5 and ByT5 in terms of data used for pretraining or model scale, we do not experiment with the models which compress the sequence length in our work.

In the context of low-resource MT, \citet{edman-etal-2022-subword} showed that character-level models can outperform subword models on the low-resource pair Xhosa--Zulu. \citet{li-etal-2021-char} showed that character-level models create higher-quality translations in synthetic low-resource settings of English$\rightarrow$\{German, Finnish\} with a corpus size of 50k parallel sentences. \citet{carrion-ponz-casacuberta-2022-effectiveness} showed that quasi-character models (subword models with a small vocabulary of size 350) produce higher-quality translations than subwords with a more standard vocabulary size of 32 thousand when data is limited for a number of European languages, finding consistent improvements across several domains.

There are two major caveats with this previous work that should be considered, however. First, previous work using character models for MT focused on training models from scratch, as this is a long-standing practice in the MT field. However, this practice could be especially harmful for low-resource languages, where the paradigm of fine-tuning a pretrained, multilingual model was shown to be effective \cite{liu-etal-2020-multilingual-denoising}. 

The second caveat is that previous evaluations of cross-lingual transfer were also limited by a relatively small model size ($<\,$70M parameters). In contrast, this work evaluates models up to 1.2B parameters. With an order of magnitude more parameters, we investigate the presence of emergent properties of larger character and subword-based NMT models, following the evidence from other generative tasks. 

Within the realm of the extensive previous work done on character models, this work serves to give an updated overview on the performance of character versus subword models for NMT. To our knowledge, we provide the first of such overviews using multilingual, transformer-based models of up to 1.2B parameters. We fine-tune a total of 162 models (varying languages, amount of training data, model size, and model type), and test on 200 languages to thoroughly compare the performance of character and subword models. We also perform an attribution analysis to gain a deeper understanding of the differences between the two granularities of character-level and subword-level, which to our knowledge, has not yet been researched.

\section{Method} \label{sect:method}

As our experiments are aimed to provide a fair comparison of character and subword pretrained models for translation, we first justify our model choice, followed by our training scheme, and lastly our choice of metric for evaluation.

\subsection{Models} \label{sect:method_models}
With many character-level models available \cite{boukkouri2020characterbert, tay2021charformer}, we opt to compare ByT5 to its subword counterpart mT5. These models are, to our knowledge, the most comparable due to their similar training setup and parameter count. 
We note that although the parameter counts between mT5 and ByT5 models are similar, \citeauthor{xue-etal-2022-byt5} opted to increase the width (i.e. hidden dimension size) of the ByT5 models to compensate for it using fewer parameters for embedding bytes. This is most noticeable in the \texttt{small} models, with 85\% of mT5's parameters being used for embeddings, compared to 0.3\% for ByT5 \cite{xue-etal-2022-byt5}. As this disparity lessens with increasing model size, we consider this difference to be a meaningful factor in explaining results correlating negatively with model size. As such, most of our experiments use the \texttt{small} (300M), \texttt{base} (582M), and \texttt{large} (1.23B) ByT5 and mT5 models, or focus only on the \texttt{large} models, where the disparity is lowest.

\subsection{Training} \label{sect:training}

We fine-tune mT5 and ByT5 models using the same prompt used in~\citet{raffel2020exploring}:
$$
\textit{Translate <S> to <T>: <src>} 
$$
\noindent where \textit{<S>} is the source language, \textit{<T>} is the target language, and \textit{<src>} is the source text. 
We primarily use WMT's NewsCommentary v16\footnote{\url{https://data.statmt.org/news-commentary/v16/}} datasets for fine-tuning. We consider 5 levels of ``resourcedness'' for fine-tuning, using \{0.4, 2, 10, 50, 250\} thousand sentence pairs. 
We also use the WMT14 German--English dataset to test higher-resource settings of 1.25 and 4.5 million sentence pairs (i.e. the entire dataset).\footnote{By default, we use NewsCommentary. Any use of WMT14 is specified.} Our hyperparameter choices for training can be found in Appendix \ref{sect:app_hyperparam}. For development and testing, we use the FLoRes-200 dataset \cite{costa2022no}.

As for training language pairs, we train on \{German, Russian\}$\leftrightarrow$English and \{Portuguese, English\}$\rightarrow$Spanish.
We choose these language pairs as they are all within NewsCommentary, guaranteeing a similar quality, and accounting for varying degrees of language similarity.

\begin{figure*}[!htp]
    \centering
    \includegraphics[width=\textwidth]{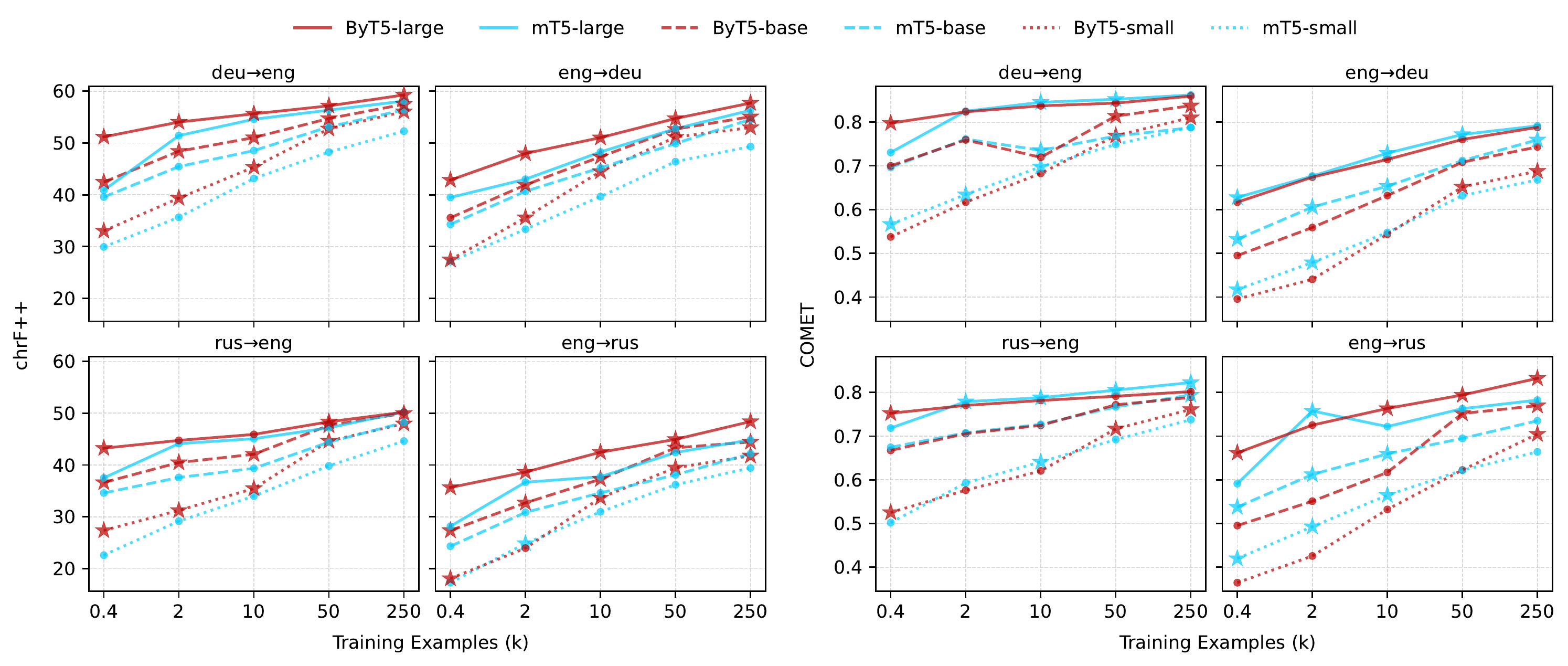}
    \caption{Translation quality using chrF++ (left) and COMET (right) of mT5 and ByT5 when fine-tuned and tested on German$\leftrightarrow$English and Russian$\leftrightarrow$English. Stars indicate a significant difference ($p<0.05$) in a paired t-test between the two respective models.}
    \label{fig:train_main}
    \vspace{-10pt}
\end{figure*}

We additionally test the models' ability to retain cross-lingual information with the FLoRes-200 dataset \cite{costa2022no}, whose wide variety of languages allows us to further isolate important language characteristics. To test the models' zero-shot capabilities, we simply swap out \textit{<S>} and \textit{<src>} for a new language, keeping the target language the same. No further training is performed, making the setting zero-shot.

\subsection{Evaluation}
We considered several translation quality metrics, % for evaluating these models, 
opting eventually for chrF++ \cite{popovic2017chrf++}, which is formulated as a combination of both character-level \textit{and} word-level F-scores, weighted to maximize correlation with human judgment. Note this differs from the original chrF, which only factors in character-level scores. Combining both word-level and character-level scores means neither ByT5 nor mT5 should be favored by chrF++.

We also considered modern neural metrics such as COMET \cite{rei2020comet,rei-etal-2022-comet}, which are known to correlate better with sentence-level human judgments \cite{freitag2022results}. However, the fact that COMET itself is a subword model, and has been mostly trained to evaluate the outputs of subword models, makes it unclear how the metric performs for character models. As such, we resort to the more transparent chrF++. Nevertheless, we provide COMET scores for the direct translation results (Section \ref{sect:dt_res}), 
%in Figure \ref{fig:train_main}
using the \texttt{wmt22-comet-da} model (i.e. COMET-22). We also include BLEU scores in Appendix \ref{app:bleu}, the trends of which are highly similar to the trends seen for chrF++.

\section{Direct Translation Results} \label{sect:dt_res}

Our direct translation setup evaluates models on the same language pairs for which they are fine-tuned. Since character models generally outperform subword models, especially when training data is scarce (as noted in Section \ref{sect:rw}), we vary the amount of fine-tuning data, confirming these findings for \{German, Russian\}$\leftrightarrow$English translation.

Varying the amount of fine-tuning data reveals the largest difference in the translation quality of character and subword models. 
Figure \ref{fig:train_main} shows that ByT5 outperforms mT5 in all resource levels according to chrF++, and is competitive with mT5 according to COMET. When resources are limited, the quality gap between ByT5 and mT5 also tends to increase.
We see that model size also plays a role, with the \texttt{large} model having the largest quality gap of up to 10 chrF++ points when only 400 sentences are available.
This goes against our assumption that, given the differences in architecture being largest between the \texttt{small} models, we would see the largest difference in performance from them (see Section \ref{sect:method_models}).

\begin{table}[!htp]
\small\centering
\begin{tabular}{lrrrrr}\toprule
&& \multicolumn{3}{c}{\# of Training Examples} \\
&&250k &1.25M &4.5M \\\midrule
\multirow{2}{*}{chrF++}& mT5 &54.72 &58.38 &61.51 \\
& ByT5 &\underline{\textbf{56.83}} &\underline{\textbf{59.78}} & \underline{\textbf{62.73}} \\ \midrule

\multirow{2}{*}{COMET}& mT5 & \textbf{0.843} & \textbf{0.857} & 0.867 \\
& ByT5 & 0.841 & 0.856 & \textbf{0.874} \\
\bottomrule
\end{tabular}
\caption{Scores on chrF++ and COMET for mT5-\texttt{large} and ByT5-\texttt{large} fine-tuned on WMT14 German$\rightarrow$English. Underlined scores are significantly better $(p < 0.05)$.}
\label{tab:train_deueng_wmt}

\end{table}

Table \ref{tab:train_deueng_wmt} shows the performance of the \texttt{large} models on the German$\rightarrow$English WMT14 dataset, accounting for higher-resource settings. 
The results according to chrF++ show that ByT5 continues to outperform mT5 by a significant margin, while COMET finds no significant difference. 
While we expect that the chrF++ performance of the two models will eventually converge given enough data, we were not able to observe it even in our highest-resource setup of 4.5M sentence pairs.

\section{Zero-shot Results} \label{sect:zs_res}
Our zero-shot evaluation examines how well the models retain information from similar languages seen in pretraining or fine-tuning or how well they generalize to unseen languages when fine-tuned on a specific language pair. As such, we consider zero-shot translation as translating from a source language that has not been fine-tuned on, using our scheme described in Section \ref{sect:training}.
This can have important implications on best practices for low-resource model training, as cross-lingual transfer learning is a common technique for obtaining a good initialization for less-resourced language pairs. 

We first look at the general translation quality across all languages in FLoRes-200, %. Within this, we note 
and study patterns in performance with respect to geography and language resourced-ness. 
We then investigate a degradation in zero-shot quality which we observe when ByT5 is trained for too long.

\begin{figure}[!htp]
    \centering
    \includegraphics[width=220pt]{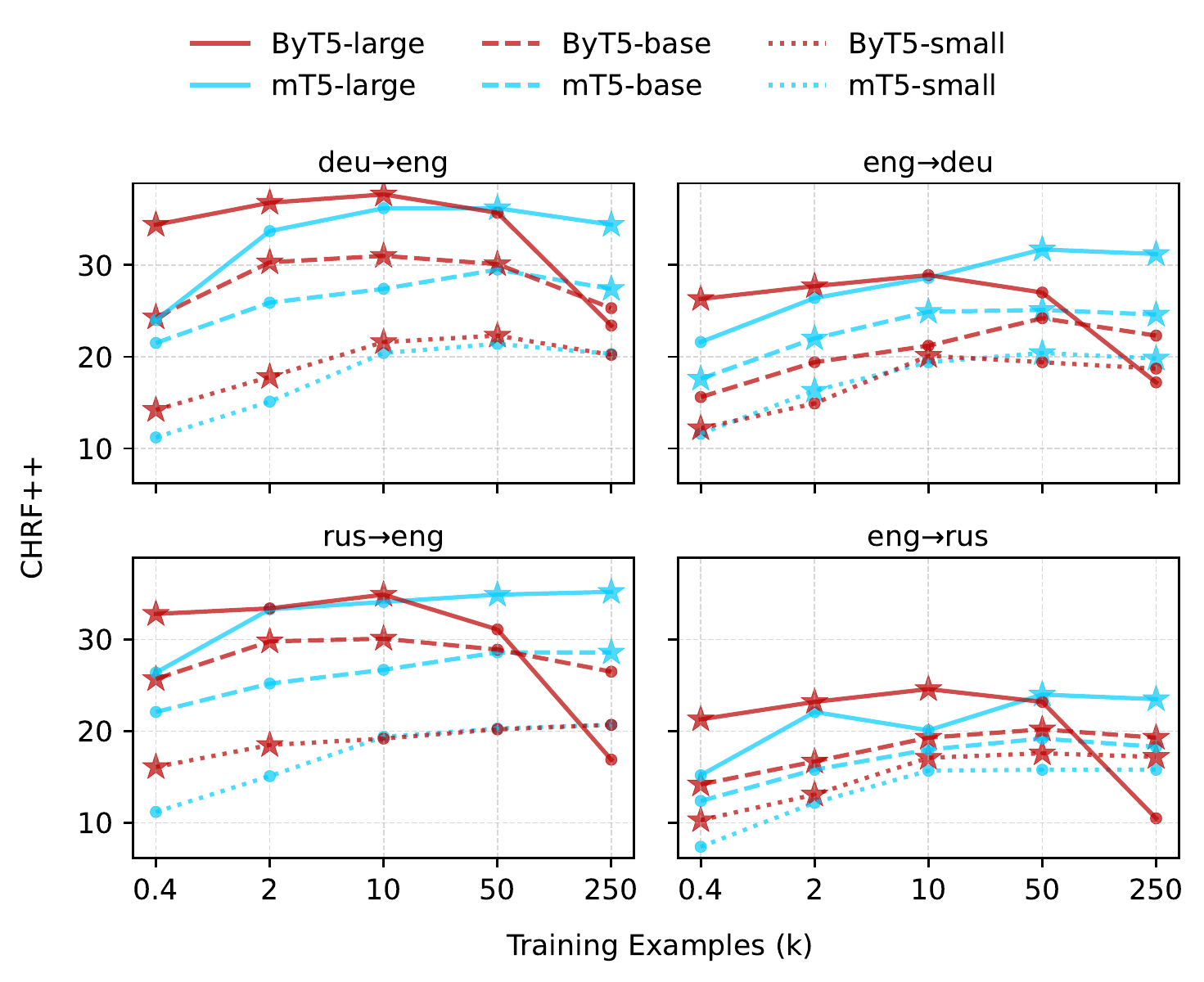}
    \caption{Zero-shot translation quality of models fine-tuned on German$\leftrightarrow$English and Russian$\leftrightarrow$English, tested on X$\rightarrow$English for all languages in FLoRes-200 (averaged results).
    Stars indicate a significant difference ($p<0.05$) in a paired t-test between the two respective models, over all language sets combined.}
    \label{fig:zs_all}
\end{figure}

\begin{figure}[!htp]
    \centering
    \includegraphics[width=220pt]{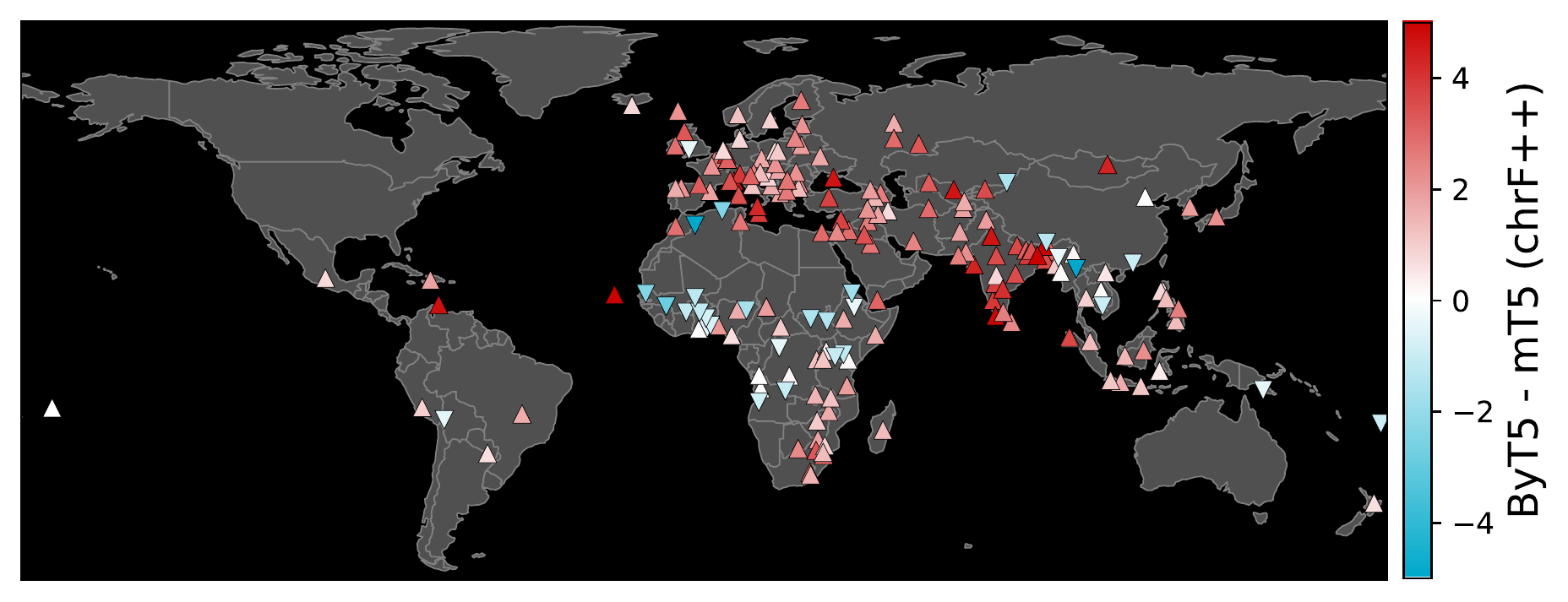}
    \caption{Map of all languages in FLoRes-200, colored according to which \texttt{large} model performed better (red ``$\bigtriangleup$'' if ByT5, cyan ``$\bigtriangledown$'' if mT5) when fine-tuned on 10k German$\rightarrow$English examples.}\label{fig:worldmap}
    \vspace{-12pt}
\end{figure}

\subsection{General Performance}
Figure \ref{fig:zs_all} shows 
the average translation quality of \{German, Russian\}$\leftrightarrow$English fine-tuned models tested on all 204 languages from the FLoRes-200 dataset (X$\rightarrow$English).

Overall, we see that ByT5 continues to outperform mT5 for lower resource scenarios. However, its zero-shot performance drastically decreases in several cases above 10k training examples, while mT5 continues to perform well up to 250k examples, with only a slight dip in performance comparatively or no dip at all.
We further investigate the large degradation of ByT5 in Section \ref{sect:degrade}, but first, we take a closer look at language-specific results on German$\rightarrow$English with 10k training examples.

In Figure \ref{fig:worldmap}, we see the performance difference of ByT5 and mT5 for each source language, plotted in their respective geographic locations.\footnote{Locations sourced from: \url{https://glottolog.org/glottolog/language}}
While ByT5 performs well overall, we notice an apparent weakness when zero-shot translating languages from West and Central Africa. Many of these languages are considered low-resource languages, however there are several languages also considered low-resource on which ByT5 performs well. 
Therefore, we break down the necessary components of a language for ByT5 to perform well next.

\subsection{Language Resourcedness}
As we have seen in Section \ref{sect:dt_res}, the amount of data used for fine-tuning plays a major role in the quality of translations. Similarly, we would expect whether the model has seen a particular language in \textit{pre-training} to play a role as well.

If we only use the presence of a language in the pretraining dataset to predict whether ByT5 outperforms mT5, we get an accuracy of 62\% on the FLoRes-200 languages, using our German$\rightarrow$English model. For our other language pairs, we get 71\%, 45\%, and 50\% for models fine-tuned on English$\rightarrow$German, Russian$\rightarrow$English,
and English$\rightarrow$Russian, respectively. So presence in pretraining alone is not a reliable predictor of performance.

However this does not tell the full story, as many low-resource languages are related to high-resource languages, which the models, particularly ByT5, are capable of exploiting.
Specifically, we categorize each language into one of three categories:
\begin{itemize}[itemsep=-3pt]
    \item High Resource: The language is in the pretraining data.
    \item Low Resource -- Related: The language is in the same language \textbf{subgrouping} (as determined by \citet{costa2022no}) and same \textbf{script} %as a high resource language 
    as a language in pre-training.
    \item Low Resource -- Unrelated: All other languages.
\end{itemize}

As an example, Acehnese (ace) is written in both Arabic and Latin scripts. For Latin, we consider it ``Low Resource -- Related'' because it is related to the high-resource Indonesian (both being Malayo-Polynesian languages), which is also written in Latin script. However for Arabic, we consider it ``Low Resource -- Unrelated'' because there are no Malayo-Polynesian languages written in Arabic script in the pretraining data.

Requiring both language subgrouping \textit{and} script to be the same performs much better as a predictor than using only one of the two. It is also intuitive: Related languages %should share a vocabulary, but if they are in different scripts then they will share much fewer embeddings, and the same is true if they are from different language subgroupings. 
typically share many similar words, but this will not result in shared embeddings if they are written in different scripts.

\begin{figure}[!htp]
    \centering
    \includegraphics[width=202pt]{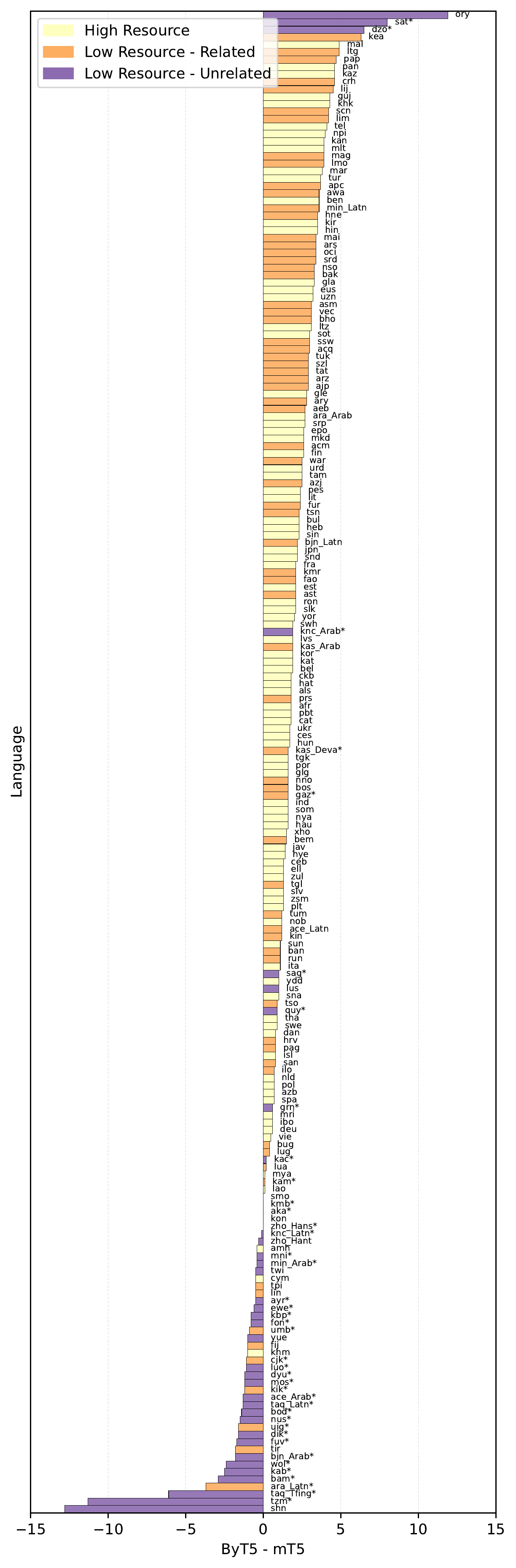}
    \caption{Average chrF++ difference for all languages in FLoRes-200, split into 3 resource categories. We use our \texttt{large} models fine-tuned on 10k German$\rightarrow$English sentences. Asterisks indicate languages which achieved a lower quality of less than 25 chrF++ for both mT5 and ByT5.}
    \label{fig:hrlrlr}
\end{figure}

In Figure \ref{fig:hrlrlr}, we see that, for the most part, languages in the first 2 categories tend to be languages where ByT5 outperforms mT5, meanwhile the reverse is true for the third category. Using this rule, we can correctly predict which model will perform better for 89\% of the languages when fine-tuning on
German$\rightarrow$English.\footnote{Similarly, we get an accuracy of 74\%, 79\%, and 81\% for our models fine-tuned on English$\rightarrow$German, Russian$\rightarrow$English, and English$\rightarrow$Russian, respectively.}

While not a perfect predictor, the efficacy of this simple rule shows there is a clear trend that ByT5 can better leverage information from other languages it has seen to inform its translation of low-resource languages. The reasoning for why mT5 performs better on unrelated low-resource languages is somewhat unclear. One possible explanation is that mT5's sparse embeddings allow for a more robust encoding for languages where false friends are more abundant, seeing as mT5 would contain less positive bias towards semantic similarity given orthographic similarity. In fact, from the perspective of a character model, ``false friends'' do not necessarily need to be words, but could also be simply character n-grams. Nevertheless, many of the languages in the third category include West and Central African languages, corresponding to the pattern we see in Figure \ref{fig:worldmap}. These languages mostly use the Latin script, increasing the likelihood of false friends with the original source language (German). It should be noted however that a majority of the languages in which mT5 performs better are also those where neither mT5 or ByT5 showed performance above 25 chrF++ (which is roughly equivalent to a BLEU score of 5). As such, we recommend further work before drawing conclusions based on these languages.

\subsection{Zero-shot Degradation}
To understand the dip in quality we see from ByT5-\texttt{large} fine-tuned on 250k examples (Figure \ref{fig:zs_all}), we begin by showing that freezing certain parts of the model can mitigate this weakness. Based on our discoveries, we then examine whether writing script is an explaining factor in how much a specific language degrades in quality.

\paragraph{Does Freezing Encoder Layers Prevent Cross-Lingual Forgetting?}
\label{sect:degrade}
We experiment with freezing various parts of the model after 2000 steps (the number of steps it takes for the models fine-tuned on 10k sentences to converge), and continuing to train on only part of the model. We proceed to test the models on their original source language (in this case German), as well as the 204 languages from the test set. Results are shown in Table \ref{tab:freeze}, including a comparison to the best-performing unfrozen models.

Here we can see that, for ByT5, freezing anywhere from the first quarter of the encoder layers up to everything except cross-attentions is effective at preventing the loss of generality that comes with extra training.
\citet{ingle-etal-2022-investigating} saw a similar pattern in the few-shot setting, where freezing the first 25\%-50\% layers of RoBERTa improved performance over freezing nothing.\footnote{\citet{gheini-etal-2021-cross} also similarly found that only freezing cross-attentions is effective in preserving translation quality, but did not test freezing only the encoder.}
This also shows that the issue is not with the variety of the training data, but specifically the number of training steps taken.

\begin{table}
\centering
\small
\begin{tabular}{lrrrrr}\toprule
&\multicolumn{2}{c}{Trained} &\multicolumn{2}{c}{Zero-shot} 
% &\multicolumn{2}{c}{(deu)} &\multicolumn{2}{c}{(Avg. 204)} 
\\\cmidrule(lr){2-3} \cmidrule(lr){4-5}
Frozen &ByT5 &mT5 &ByT5 &mT5 \\\midrule
Embeds &59.1 &\textbf{58.7} &26.1 &\textbf{36.3} \\
Enc$_{12.5\%}$ &59.0 &58.1 &30.7 &34.8 \\
Enc$_{25\%}$ &58.7 & 57.9 &37.2 & 35.5 \\
Enc$_{50\%}$ &57.2 &32.7 &\textbf{38.1} &19.1 \\
Enc$_{100\%}$ &57.3 &31.1 &37.6 &19.3 \\
$\neg$ X-Attn &57.4 &30.5 &37.9 &17.9 \\ \midrule
10k &55.6 &55.1 &37.7 &36.2 \\
250k &\textbf{59.3} &58.1 &23.4 &34.4 \\
\bottomrule
\end{tabular}
\caption{chrF++ on Flores-200 for trained language pair (``Trained'', i.e. German) and all remaining languages (``Zero-shot'', averaged) using \texttt{large} models fine-tuned on 250k German$\rightarrow$English examples. \textbf{Top:} Performances by freezing some model components (``Enc$_{n\%}$'': freeze the first $n\%$ of encoder layers; ``$\neg$ X-Attn'': freeze all parameters \textit{except} cross-attentions.) \textbf{Bottom:} Performance of unfrozen models trained for 10k and 250k steps (i.e. the step counts with the best performing models for zero-shot and trained settings).}\label{tab:freeze}
\vspace{-10pt}
\end{table}

As we see a relatively steep incline in zero-shot translation quality from 0 to 25\% of the encoder layers frozen, and then a relatively stable quality afterwards, it seems the majority of language-specific operations are occurring in the first 25\% of the layers of the encoder in ByT5.  On the other hand, mT5 appears to exhibit this language abstraction solely based on its sparser embeddings, as it does not suffer the same level of zero-shot quality decrease when not frozen. However, when freezing layers of mT5 beyond the embeddings, we see a steep \textit{decrease} in translation quality both on the trained language and zero-shot languages. This implies that mT5 needs to adjust a much larger portion of its parameters in order to translate with higher quality, whereas ByT5 only needs to adjust at most its cross attentions. 

In light of minimal gains from further training on zero-shot, early stopping using a left-out set of low-resource language examples seems ideal.

\paragraph{Does Writing Script Affect Degradation?}
The fact that freezing the encoder preserves the quality on zero-shot languages narrows the failure point of ByT5 down to the source side. Since a major difference between character and subword models is their respective dense and sparse token embeddings, we examine the impact of each language's writing script on the degradation of translation quality. We expect that, for example, with our German$\rightarrow$English models, other Latin-scripted languages will be more severely impacted, due to their high degree of embedding overlap. We expect those embeddings (and subsequent encodings) to become specialized to German only, meanwhile non-Latin encodings should remain relatively unchanged.

\begin{figure}[!htp]
    \centering
    \includegraphics[width=220pt]{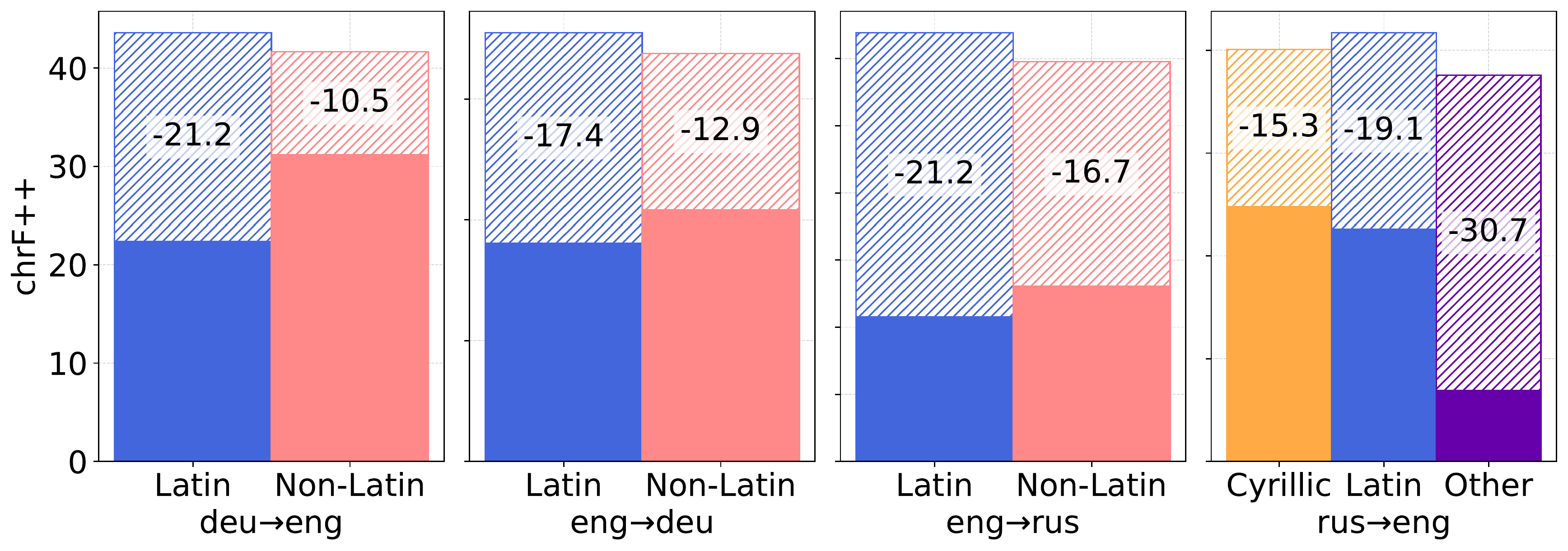}
    \caption{Effect of language script on chrF++ degradation when fine-tuning for longer (250k (solid) vs. 10k (solid + striped) examples).}
    \label{fig:degrade}
    \vspace{-10pt}
\end{figure}

Figure \ref{fig:degrade} shows the degradation factor of Latin versus Non-Latin-scripted languages for our ByT5-\texttt{large} English$\leftrightarrow$German and English$\rightarrow$Russian models and Latin, Cyrillic, and all other multi-byte scripts for our Russian$\rightarrow$English models.\footnote{We remove all languages which have a score of 25 chrF++ or less on 10k, since these languages score so low that degradation from more training is not relevant.} 

For the first 3 subfigures, we can see that Latin-scripted languages degrade more than the  non-Latin languages. This is expected, seeing as ByT5 shares embeddings (and presumably encodings) for all Latin languages. So, for example, when it is specifying for German$\rightarrow$English, any Latin-scripted languages unrelated to German should be particularly negatively affected. Meanwhile other scripts share fewer bytes, and thus would be less affected.

For Russian$\rightarrow$English, we do not see the same trend with Cyrillic, but rather other scripts are heavily affected. We suspect this is for two reasons. Firstly, the Cyrillic languages are much more concentrated around Russia than Latin languages are around Europe, so the Cyrillic languages tend to be either more closely related to, and partially mutually intelligible with, Russian (e.g. Ukrainian), or have several loan words from Russian (e.g. Kazakh). Meanwhile the other non-Latin, non-Cyrillic scripts are heavily affected because they are also multi-byte scripts. These scripts share a large portion of bytes with Cyrillic and each other due to how UTF-8 encodes multi-byte characters. For example, the ``\foreignlanguage{russian}{I}'' in Cyrillic is encoded in bytes as [208, 152], while the Devanagari ``{\dn G}'' is [224, 164, 152]. % घ
Both share the final byte as both are the 25th character within their respective code blocks. 

As such, we find that multi-byte scripts may suffer from zero-shot translation degradation due to vocabulary overlap with the original source-side script, assuming this source-side script is another multi-byte script.

Since we observe a loss of generality only when at least 50k sentences are used for fine-tuning, the following sections mostly focus on results from models using between 0.4k and 10k fine-tuning examples.

\section{Attribution Analysis} \label{sect:analysis}
In translation studies, words are widely recognized as the minimal unit of translation~\citep{hatim2004translation}. 
It is therefore natural to assume that a source-to-target word or subword mapping might be easier to create than a character-level one, except perhaps for closely related languages where many words translate into orthographically similar cognate words. 
Therefore, it seems that (sub)word models are naturally more suited towards machine translation, but our results from the previous sections appear to contradict that.
This naturally leads us to investigate whether and how character models can learn to incorporate word-level information to improve their translation capabilities.

A common way to quantify the influence of inputs in driving model predictions is by means of \textit{feature attribution methods}~\citep{madsen-etal-2022-posthoc}, which have previously been applied in the MT domain to highlight word alignments, coreference resolution capabilities, and model training dynamics (\citealp{ding-etal-2019-saliency,he-etal-2019-towards,voita-etal-2021-language} \textit{inter alia}). 
For our analysis, we study source versus target character contributions in ByT5 models using gradients~\citep{simonyan2013deep}, which were recently shown to be more faithful than other advanced approaches for language tasks~\citep{bastings-etal-2022-will}. 
One can take gradients with respect to the probability of a predicted token and propagate them back to the model's inputs in order to quantify the importance of input tokens to the resulting prediction. 
Our analysis is inspired by~\citet{voita-etal-2021-analyzing}, where a similar analysis was conducted on subword MT models to highlight a ``language modeling regime'' where high importance is given to the target prefix while disregarding source tokens. 

In our study, we compute gradients for each input token (i.e. a character) with respect to the next token's prediction probability at every generation step using the Inseq library~\citep{inseq}. For every generated token, gradient vectors are aggregated at the character level using the L2 norm of the gradient vector to gauge the influence of the source and target context on the model's predictions.\footnote{We normalize the total source + target contribution to 1 to obtain relative contributions, similar to~\citet{voita-etal-2021-analyzing}.}

We verify our hypothesis that character-level NMT models might implicitly operate at a word level by comparing source-side and target-side contributions across different model sizes and language pairs. In this context, a \textit{character-by-character} translation would be characterized by relatively uniform source attributions across all generated characters, while a \textit{word-by-word} translation would imply a higher source contribution of the characters marking the beginning of a new word, followed by a shift of importance towards the target prefix indicating that completing the word requires limited access to source-side information. Figure~\ref{fig:attr_example} provides an example of source and target contributions, showing a marked increase in source importance at the beginning of each word.

\begin{figure}
    \centering
    \includegraphics[width=0.48\textwidth]{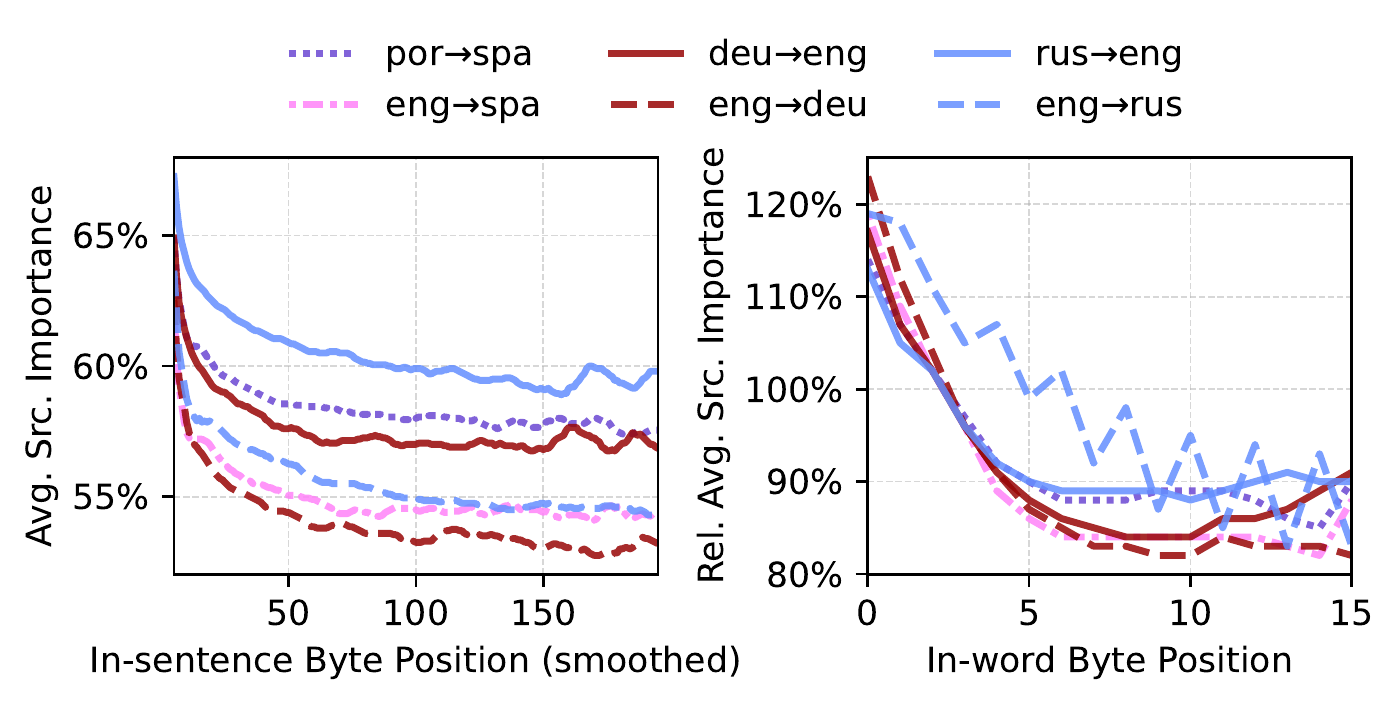}
    \caption{\textbf{Left:} Average ByT5-\texttt{large} source importance across byte positions within a sentence. Values are smoothed using a 10-point rolling window. \textbf{Right:} Average ByT5-\texttt{large} source importance across bytes within a word, normalized by sentence position (left plot) to account for the effect of decreasing source importance.
    }
    \label{fig:attr_pos}
    \vspace{-10pt}
\end{figure}

\paragraph{How is Word-level Information Used in Character Models?} Figure \ref{fig:attr_pos} (left) confirms the decline in source importance throughout generation observed by~\citet{voita-etal-2021-analyzing} for several ByT5 models. 
In the left plot, we compute the average source importance with respect to the character's position in the sentence. While progressing through the generation, we observe that the model relies less on the source side and more on the target side for its next-byte predictions. This aligns with \citet{voita-etal-2021-analyzing}'s findings for subword-level MT models, with the intuitive explanation that a longer generated context can provide more cues for the model to produce a coherent output without the need to attend to source-side tokens.

In the right plot, instead, we compute the \textit{relative} average source importance for byte positions within words.
Since bytes occurring later in a word also occur later in a sentence, we would expect source importance to drop later in a word as a consequence of the trend seen in the left plot.
To disentangle the effect of the in-word byte position from the sentence-level one, we normalize the source importance scores by their average for their corresponding position in the sentence (i.e., we divide every in-word score by the respective in-sentence average from the left plot). As a result, we would expect stable relative source importance close to 100\% across all in-word byte positions if the position of characters inside the word did not affect source importance.
Instead, we observe a rapid decline of source importance within the word across all languages, implying an increase in target-side importance for non-word-initial generated characters. 
Upon manual inspection of the importance patterns, we note that the target-side importance is focused mainly on the previous characters belonging to the same word, suggesting the usage of word-level information to facilitate translation.

Interestingly, for English$\rightarrow$Russian we observe an oscillatory importance pattern peaking on even-numbered positions. The observed trend reflects the UTF-8 Cyrillic encoding using 2 bytes per character,\footnote{The first identifies the character subset, and the second specifies the character.} with peaks at first-byte boundaries pointing to the model's ability to discriminate characters and bytes in languages using multi-byte characters.
We also note that relative source attributions seem to converge to 100\% around the third character for Latin script languages and around the sixth in-word character for Russian. These results indicate that similar importance patterns might emerge across different languages in character models despite separate fine-tuning procedures.

\begin{figure}
    \centering
    \includegraphics[width=0.48\textwidth]{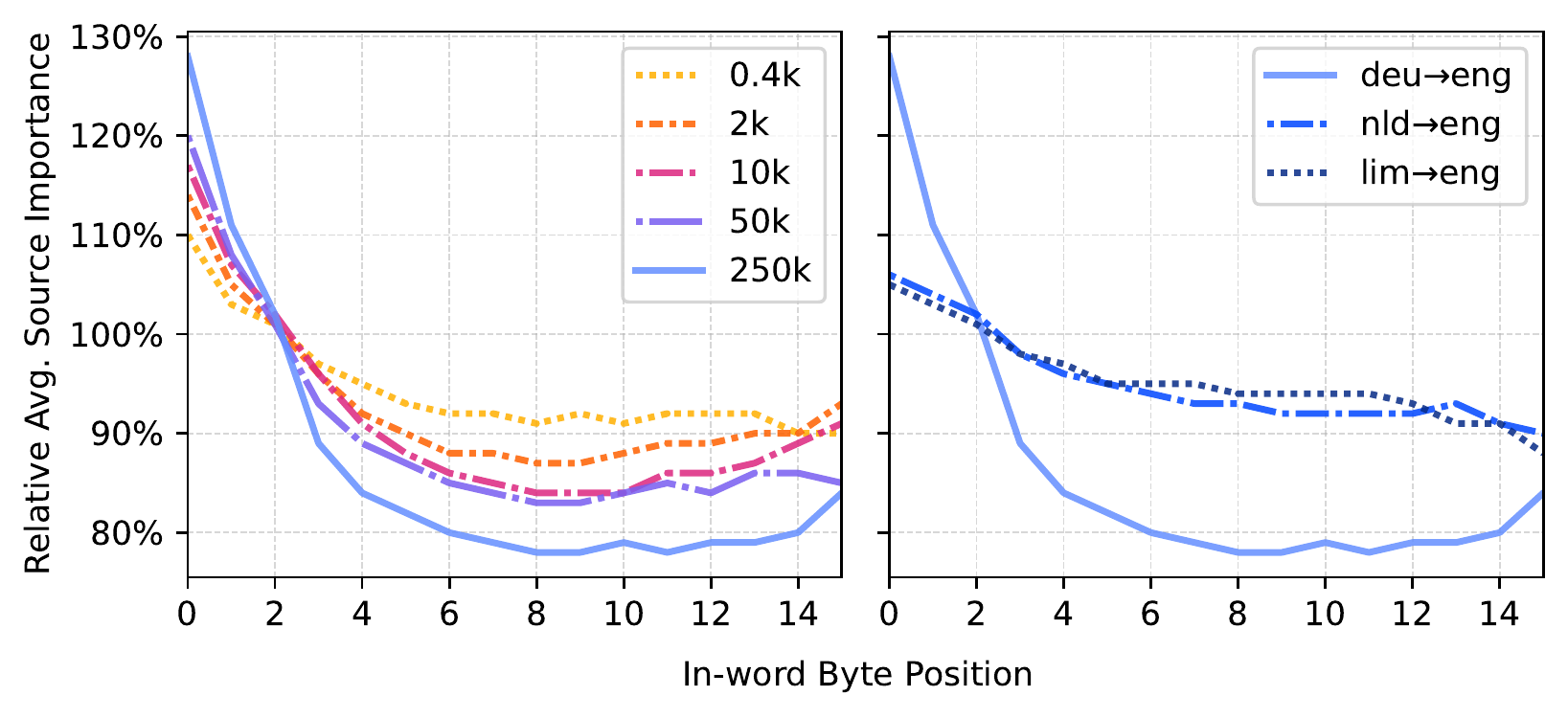}
    \caption{\textbf{Left:} Source contribution across word's characters for a ByT5-\texttt{large} fine-tuned on deu$\rightarrow$eng translation with different \# of fine-tuning examples (0.4k to 250k). \textbf{Right:} Source contributions for same-language (deu$\rightarrow$eng) and zero-shot cross-lingual translation using the same 250k ex. model.}
    \label{fig:attr_pos_per_step}
    \vspace{-8pt}
\end{figure}

\paragraph{How does Resourcedness Affect Word-level Modeling?} In Figure~\ref{fig:attr_pos_per_step}, we examine how the relative source importance changes when more training data becomes available, and its difference between trained and zero-shot languages. The left plot shows that a larger amount of training examples contributes to sharpening the source contribution around the first few bytes in a word while decreasing source importance for the following bytes. These results support our intuition that memorized co-occurrences between source and target words, which are captured more easily when given more training examples, enable confident generation with less reliance on the source text. From the right plot, we also note how this trend does not automatically extend to related languages when performing zero-shot translation, with source importance patterns for \{Dutch, Limburgish\}$\rightarrow$English translation using a German$\rightarrow$English model presenting a smooth trend comparable to the 0.4k German$\rightarrow$English one in the left plot. This suggests that memorized co-occurrences lowering source-side dependence do not generalize to related languages even when translation quality does.

\section{Effect of Word Similarity and Frequency}\label{sec:cognates}

Character models have previously been shown to have a more robust vocabulary and are thus able to form useful biases for translating orthographically similar words, as well as rare words \cite{lee-etal-2017-fully}. 
We revisit these findings to see if they hold with larger, multilingually-pretrained Transformers. 

\paragraph{Can Character Models Exploit Word Similarity?}
We proceed to investigate whether character models can operate on character level when desirable. To this end, we focus our analysis on orthographically similar words (OSWs), 
starting from the assumption that a character model can easily exploit their similarity by learning to copy matching substrings.
We start by assessing whether character models show improved performances for OSW translation. We use \texttt{AWESoME}~\citep{dou-neubig-2021-word} to align source, reference, and hypothesis sentences and calculate word-level translation accuracy. Our definition of OSWs is based on the inverse normalized-Levenshtein distance of the source and reference words, varying the threshold from 0 to 100\%, e.g. German \textit{gesundheit} and Dutch \textit{gezondheid} have a similarity of 70\% since they differ in three letters.

\begin{figure}[!htp]
    \centering
    \includegraphics[width=0.48\textwidth]{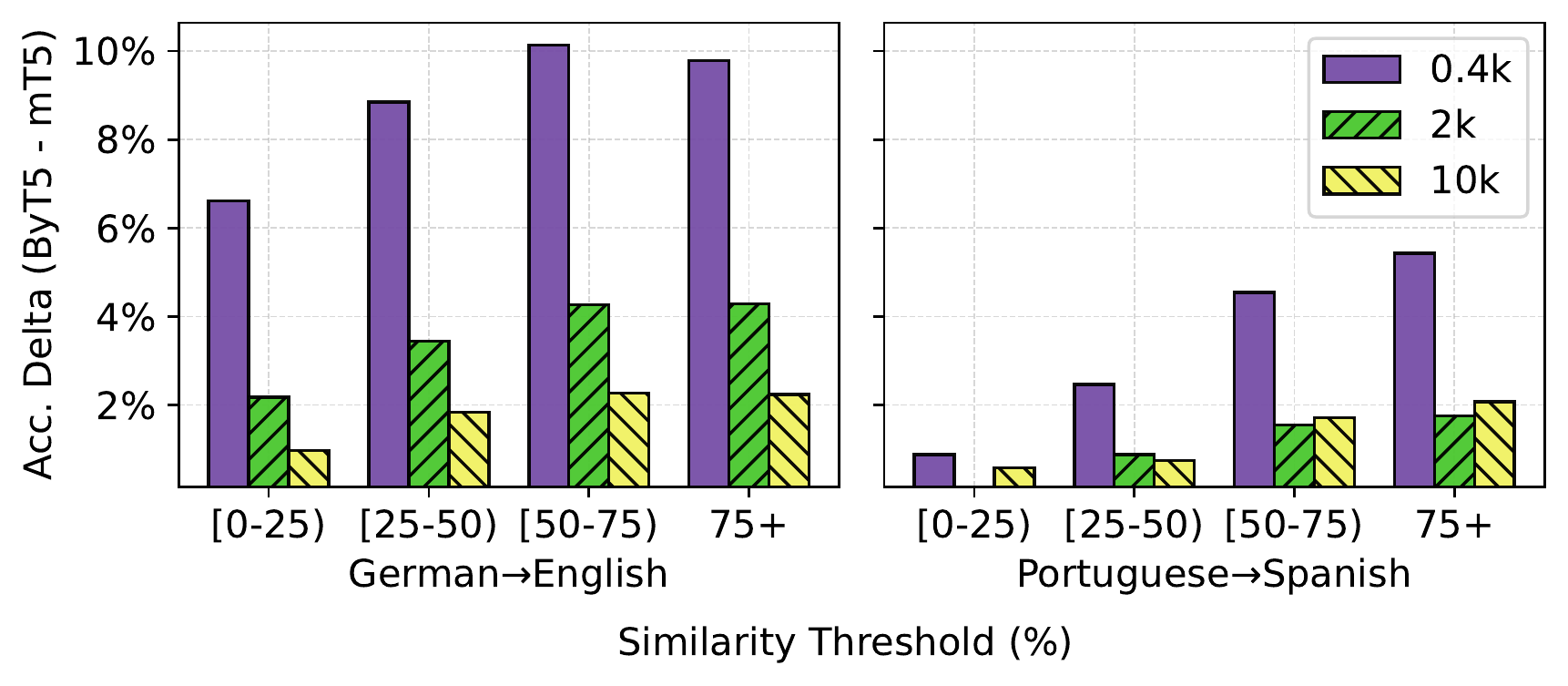}
    \caption{Word-level accuracy deltas for \texttt{large} models at different orthographic similarity levels, trained on different numbers of examples (0.4k, 2k, 10k).}
    \label{fig:cognate_acc}
\end{figure}

Figure \ref{fig:cognate_acc} shows the accuracy difference for the \texttt{large} ByT5 and mT5 models trained for German$\rightarrow$English and Portuguese$\rightarrow$Spanish on words grouped at different levels of orthographic similarity. We observe that, as words become more similar, the accuracy delta also increases in favor of ByT5, especially when less fine-tuning data is used. These results indicate that character models can learn OSW translations more rapidly and effectively than their subword counterparts. 

\begin{figure}[!htp]
    \centering
    \includegraphics[width=0.48\textwidth]{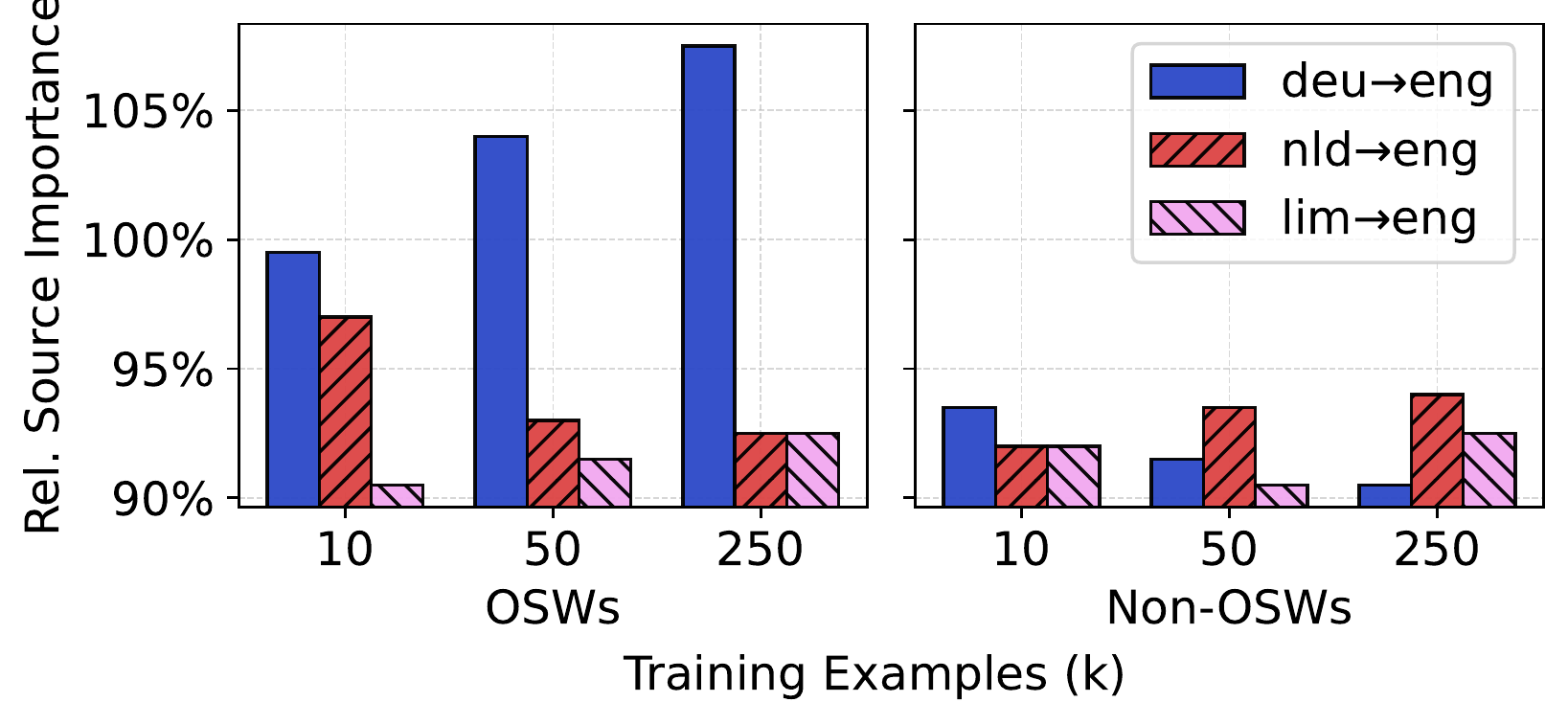}
    \caption{Relative source importance for ByT5 \texttt{large} fine-tuned on German$\rightarrow$English translation for OWSs (>70\% sim.) and non-OSWs (<30\% sim.) in same-language and zero-shot settings, given different numbers of fine-tuning examples.}
    \label{fig:cognate_attribution}
    \vspace{-12pt}
\end{figure}

We proceed by examining the source contributions of OSWs and non-OSWs using ByT5, and how those change according to the number of examples used for fine-tuning. 
We restrict our analysis to >70\% similarity for OSWs, and <30\% similarity for non-OSWs. We consider only the source importance over characters belonging to the same word, in order to investigate the presence of copying patterns for OSWs.
As shown in Figure~\ref{fig:cognate_attribution}, the importance of source text grows with the amount of fine-tuning data when translating OSWs in the German$\rightarrow$English setting (same direction as training), suggesting an improved ability of the model in copying information from the source. Conversely, we observe a declining trend for the zero-shot Dutch$\rightarrow$English direction, converging to the source importance of Limburgish, which was not present in the model's pretraining data. This could indicate that the Dutch knowledge acquired during pretraining is progressively lost when fine-tuning on increasingly more German data, supporting the results of Figure~\ref{fig:zs_all}. For non-OSWs, we find a relative source importance of $94\pm4$\% across all three directions and all amounts of training examples, indicating that the observed copying behavior is restricted to highly similar words only. 

To further confirm this copying behavior is indeed distinct between a subword and character model, we create a synthetic \textit{control test set} by replacing all proper noun pairs (word pairs tagged as proper nouns in both source and reference target sentences\footnote{POS tags obtained via NLTK's \texttt{PerceptronTagger}. Roughly 6\% of the test set's words are tagged as proper nouns.}) with random strings of Latin alphabet characters matching the original words' length. This is done for our German$\rightarrow$English \texttt{large} models, modifying the German$\rightarrow$English test set from FLoRes200. Table \ref{tab:nonsense} reports subword and character models' accuracies in copying the random strings from the source input into the generated output. 

\begin{table}[!htp]\centering
\small
\begin{tabular}{p{27pt}p{15pt}ccccc}\toprule
& & \multicolumn{5}{c}{\# of Training Examples} \\
& &400 &2k &10k &50k &250k \\\midrule
\multirow{2}{*}{Acc (\%)} &mT5 &25 &57 &69 &61 &71 \\
&ByT5 &\textbf{91} &\textbf{90} &\textbf{90} &\textbf{89} &\textbf{92} \\ \midrule
\multirow{2}{*}{chrF++} &mT5 &33 &49 &53 &55 &58 \\
&ByT5 &\textbf{51} &\textbf{54} &\textbf{56} &\textbf{57} &\textbf{59} \\ \midrule
chrF++ &mT5 &41 &51 &55 &56 &58 \\
(orig.) &ByT5 &\textbf{51} &\textbf{54} &\textbf{56} &\textbf{57} &\textbf{59} \\

\bottomrule
\end{tabular}
\caption{Per-word copying accuracies and sentence-level chrF++ scores of \texttt{large} German$\rightarrow$English models on the control test set (top, middle), and chrF++ scores on the original test set (bottom).}\label{tab:nonsense}
\end{table}

We observe that for the subword-level mT5 model, the copying mechanism is learned progressively during fine-tuning. Meanwhile, the ByT5 model instead achieves superior copying performances with very little supervision, achieving 91\% accuracy with only 400 training examples. We also observe that mT5's performance on chrF++ improves with increasing examples, while ByT5's performance remains relatively unaffected. Comparing this to the chrF++ scores on the original test set, we see that ByT5's scores are unchanged, while mT5's scores drop by up to 8 points when the test set is modified.

While our control set only targets proper nouns, this copying behavior may be far more pervasive, as it could be present in any loan words, regardless of part-of-speech. The large difference in scores, particularly on the models fine-tuned on only 400 examples, seems to show a major difference in the default behavior of the two models.

\paragraph{Can Character Models Translate Rare Words More Accurately?}
Prior work has shown character models achieving better translation accuracy of rare words \cite{lee-etal-2017-fully,singh2017handling}, so we expect the same to be true for large, pretrained, Transformers as well. We adopt the same word-level translation accuracy method as earlier in the section, but instead binning words based on their frequency in the English fine-tuning data. We evaluate word-level accuracy of ByT5 versus mT5-\texttt{large} when fine-tuned on 10k German$\rightarrow$English sentence pairs. We additionally distinguish between accuracy using the original source language and accuracy using zero-shot languages.

\begin{figure}[!htp]
    \centering
    \includegraphics[width=0.48\textwidth]{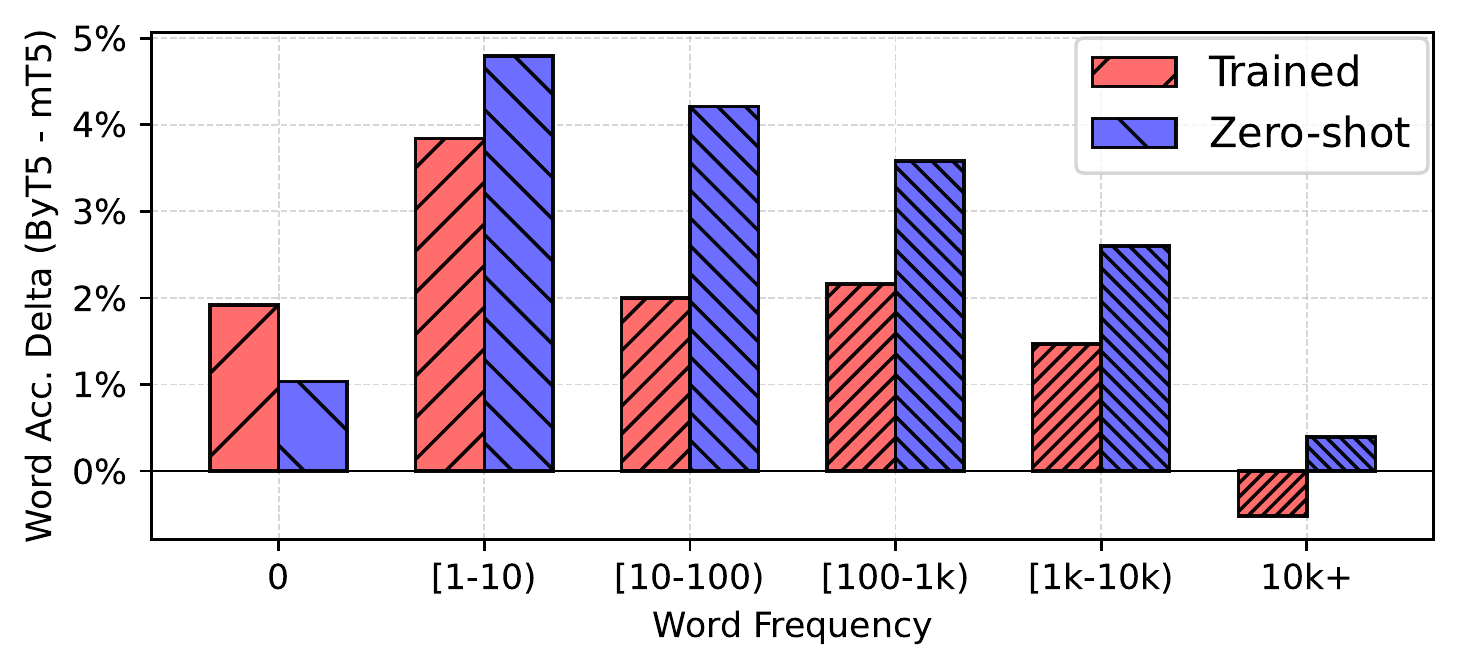}
    \caption{Word-level accuracy delta on the original source (German) and zero-shot (X$\rightarrow$English, averaged) languages, using \texttt{large} German$\rightarrow$English models with 10k training pairs. Bins based on frequencies in English fine-tuning data. }
    \label{fig:word_freq}
\end{figure}

As shown in Figure \ref{fig:word_freq},
ByT5 has a higher per-word translation accuracy across all frequencies for the zero-shot languages, and all but the most frequent words for German. Generally, as the frequency of the word increases, model disparity decreases. This is expected based on previous work \cite{singh2017handling}. The disparity is higher for zero-shot languages overall, which indicates a higher level of robustness to changes in vocabulary from ByT5, a phenomenon also previously observed \cite{belinkov2017synthetic}.

\section{Efficiency} \label{sect:time}
Although we have shown that the translation quality of character models is competitive or better than subword models, another important aspect is the efficiency of character models. 
While the lack of efficiency of character-level Transformer models is well-documented \cite{libovicky-etal-2022-dont,tay2021charformer,edman-etal-2022-subword}, we have yet to see their efficiency on larger (300M+ parameters) Transformers for NMT, and so we provide our findings for completeness. 
We report model training and inference times for 1 NVIDIA V100 32GB GPU in Table \ref{tab:times}.

\begin{table}[!htp]
\centering\small
\begin{tabular}{lrrrr}\toprule
& \multicolumn{2}{c}{Training} & Inference\\
Model & samples/s & epochs & samples/s \\\midrule
mT5-\texttt{small} &\textbf{2.50} &38.11 &\textbf{52.91} \\
ByT5-\texttt{small} &0.43 &\textbf{29.24} &8.90 \\
\midrule
mT5-\texttt{base} &\textbf{1.15} &22.87 &\textbf{20.77} \\
ByT5-\texttt{base} &0.24 &\textbf{18.16} &3.96 \\
\midrule
mT5-\texttt{large} &\textbf{0.48} &19.58 &\textbf{6.23} \\
ByT5-\texttt{large} &0.12 &\textbf{16.25} &1.17 \\
\bottomrule
\end{tabular}
\caption{The training and inference speeds for German$\rightarrow$English experiments. Epochs reported are using the models fine-tuned on 10 thousand pairs, using early stopping. The best result per model size and column is shown in bold.}\label{tab:times}
\end{table}

Both training and inference speeds in samples per second are considerably slower for the character models (4-5 times slower for training, 5-6 times slower for inference). 
The number of epochs needed to converge is lower for character models, but not enough to counteract the slowness of training. 

The slower speed comes largely from the increase in sequence length. While we tried to balance the size of the batches such that each model sees the same amount of text per batch, achieving this required the character models to accumulate gradients for 4 times as many iterations as the subword models. 
Thus, if training or inference speed is a concern, subword models are likely the superior choice, particularly for high-resource languages. In the low-resource setting, there is a significant trade-off between accuracy and speed when choosing between character and subword models. 

It should be noted that there are several alternatives to vanilla character or byte-level models which offer a speedup, typically by compressing the sequence length before the bulk of the computation is done in the Transformer (\citealp{boukkouri2020characterbert, clark-etal-2022-canine, yu2023megabyte}, \textit{inter alia}). None of these methods claim faster processing over a standard subword-level model, however, so there would still be a trade-off between accuracy and speed when applying these methods, albeit to a lesser extent.

\section{Conclusion} \label{sect:conclusion}
Subword-level models are currently the dominant paradigm for machine translation. However, this work showed that character models could produce competitive or even superior results in many circumstances. First, character models obtain an overall better translation quality on trained language pairs. Moreover, we highlighted how the gain in translation quality from using character models is particularly marked when fine-tuning data is scarce. Finally, we showed how character models have superior cross-lingual transferability, especially with languages seen in pretraining, or those that are similar to a language seen in pretraining. 

Following the results of our analyses, we can attribute this superior quality to a character model's ability to implicitly account for the appropriate input granularity given the context, translating at times in a word-by-word fashion, or character-by-character when needed. This ultimately results in better accuracy when translating orthographically similar and rare words.

The quality increase is however not without a trade-off. Indeed, character models are at least 4 times slower in training and inference, making them suboptimal for many real-world situations. We posit that further work into speeding up generation models (\citealp{stern-etal-2018-blockwise,edman-etal-2022-subword,santilli-etal-2023-accelerating}, \textit{inter alia}), as well as more thorough evaluations of neural metrics for character models, would greatly benefit this area of research. 
Nevertheless, we can conclude that in less time-sensitive, or low-resource settings, character-level translations \textit{are} worth the wait.

\section*{Acknowledgements}
We thank the TACL reviewers and action editor for their helpful feedback. We thank the Center for Information Technology of the University of Groningen for their support and for providing access to the Peregrine and Hábrók high performance computing clusters. We thank the authors of \citet{xue-etal-2022-byt5} for providing more detailed results which helped formulate the trajectory of this work.
Sarti and Bisazza are funded by the Dutch Research Council (NWO project InDeep NWA.1292.19.399).

\bibliography{anthology,custom}
\bibliographystyle{acl_natbib}

\appendix

\section{Hyperparameters} \label{sect:app_hyperparam}
We train models using the AdaFactor~\citep{shazeer2018adafactor} optimizer with 4000 warmup batches and a final constant learning rate of 1e-4. We batch by \# of tokens (20k for ByT5, 5k  for mT5) to provide each model with a roughly equivalent amount of context, as we estimate roughly 4 bytes per mT5 token during initial testing. We use early stopping with a patience of 5, based on a set number of steps, varying with the amount of data used. The step sizes were \{50, 100, 250, 500, 1000\} batches for \{0.4k, 2k, 10k, 50k, 250k\} total training examples, respectively. Similarly, we set a maximum number of epochs to \{6250, 1250, 250, 50, 10\}.
All of our experiments on WMT14 data used the same step sizes and max epochs as our experiments using 250k training examples. 

\section{Additional scores}\label{app:bleu}
\begin{figure}[!htp]
    \centering
    \includegraphics[width=220pt]{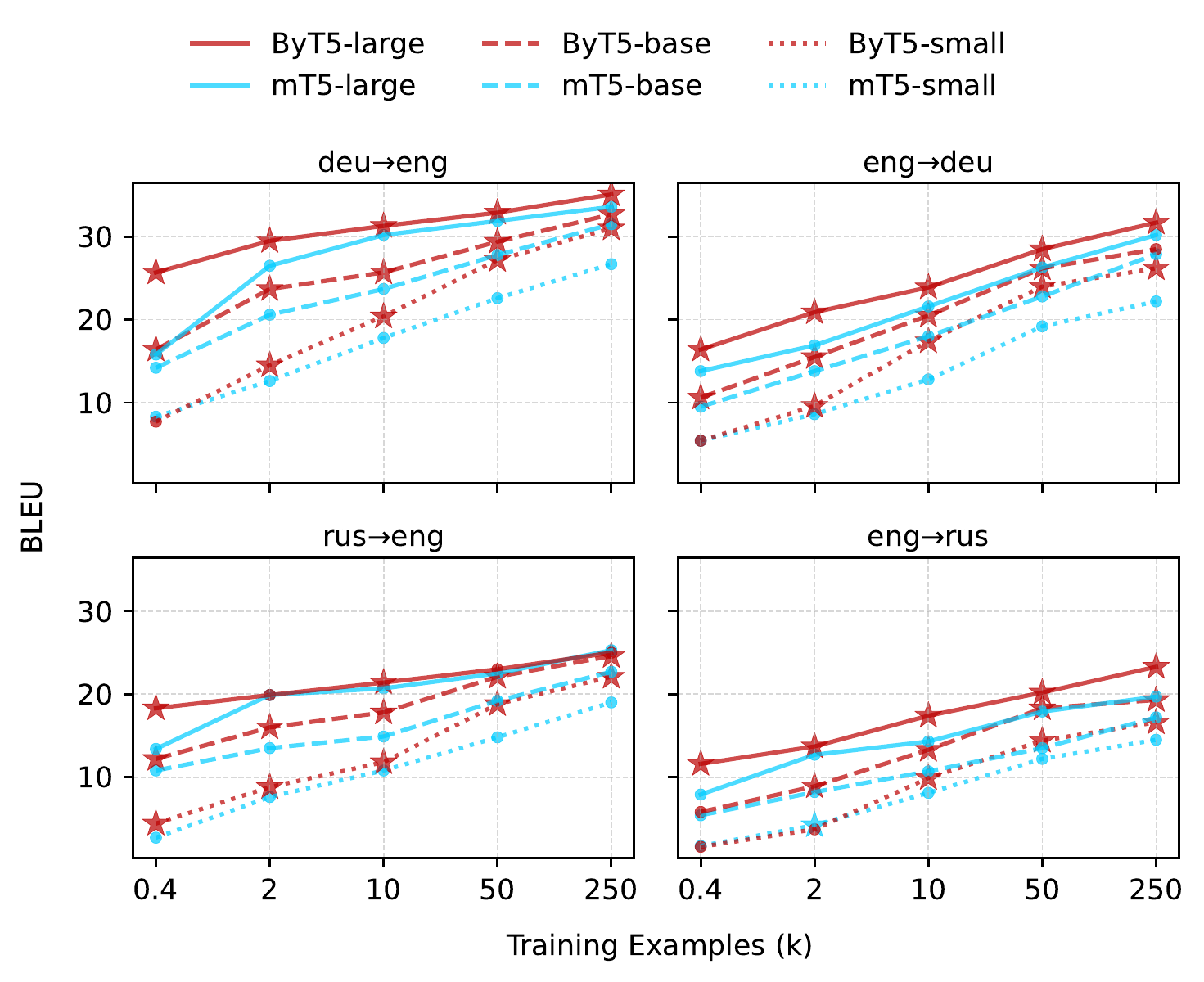}
    \caption{BLEU scores of mT5 and ByT5 on German$\leftrightarrow$English and Russian$\leftrightarrow$English. Stars indicate significant difference ($p<0.05$) in a paired t-test between the two respective models.}
    \label{fig:main_bleu}
    \vspace{-10pt}
\end{figure}

Figure \ref{fig:main_bleu} shows our main results in terms of BLEU. We can see that the trends of the BLEU scores are very similar to those of the chrF++ scores, and generally favor ByT5. 

We also provide Table \ref{tab:all_langs_deen}, which shows the raw chrF++ scores that are used to create Figures \ref{fig:worldmap} and \ref{fig:hrlrlr}.

All individual scores used to create the figures in this work (which totals over 24 thousand values), as well as the models weights and models outputs
, are provided.\footnote{\url{https://github.com/Leukas/CharLevelMT}}
% will be provided upon publication.

\begin{landscape}
\begin{table}[!htp]\centering
\tiny
\begin{tabular}{lrrrrrrrrrrrrrrrr}\toprule
Language &Script &mT5 &ByT5 &Language &Script &mT5 &ByT5 &Language &Script &mT5 &ByT5 &Language &Script &mT5 &ByT5 \\\cmidrule(lr){1-4} \cmidrule(lr){5-8} \cmidrule(lr){9-12} \cmidrule(lr){13-16}
Acehnese &Arabic &13.7 &12.4 &Faroese &Latin &43.0 &45.1 &Lombard &Latin &42.2 &46.1 &Samoan &Latin &40.6 &40.6 \\
Acehnese &Latin &32.2 &33.4 &Fijian &Latin &27.5 &26.5 &Latgalian &Latin &32.4 &37.3 &Shona &Latin &35.4 &36.4 \\
Mesopotamian Arabic &Arabic &40.4 &43.0 &Finnish &Latin &43.2 &45.8 &Luxembourgish &Latin &52.2 &55.3 &Sindhi &Arabic &37.1 &39.3 \\
Ta'izzi-Adeni Arabic &Arabic &41.3 &44.3 &Fon &Latin &13.0 &12.2 &Luba-Kasai &Latin &25.6 &25.8 &Somali &Latin &35.2 &36.8 \\
Tunisian Arabic &Arabic &37.2 &39.9 &French &Latin &53.8 &55.9 &Ganda &Latin &27.3 &27.7 &Southern Sotho &Latin &39.4 &42.4 \\
Afrikaans &Latin &60.7 &62.5 &Friulian &Latin &44.5 &46.9 &Luo &Latin &20.8 &19.7 &Spanish &Latin &48.1 &48.8 \\
South Levantine Arabic &Arabic &42.2 &45.1 &Nigerian Fulfulde &Latin &19.1 &17.4 &Mizo &Latin &24.3 &25.3 &Tosk Albanian &Latin &49.8 &51.6 \\
Akan &Latin &24.8 &24.8 &Scottish Gaelic &Latin &39.5 &42.8 &Standard Latvian &Latin &45.4 &47.3 &Sardinian &Latin &47.9 &51.3 \\
Amharic &Ge'ez &35.0 &34.6 &Irish &Latin &45.3 &48.1 &Magahi &Devanagari &39.2 &43.1 &Serbian &Cyrillic &50.1 &52.8 \\
North Levantine Arabic &Arabic &39.7 &43.4 &Galician &Latin &52.6 &54.2 &Maithili &Devanagari &37.8 &41.2 &Swati &Latin &32.4 &35.4 \\
Modern Standard Arabic &Arabic &43.7 &46.4 &Guarani &Latin &22.0 &22.6 &Malayalam &Malayalam &36.6 &41.5 &Sundanese &Latin &44.9 &46.0 \\
Modern Standard Arabic &Latin &21.8 &18.1 &Gujarati &Gujarati &40.1 &44.4 &Marathi &Devanagari &37.6 &41.4 &Swedish &Latin &56.2 &57.1 \\
Najdi Arabic &Arabic &42.7 &46.1 &Haitian Creole &Latin &46.4 &48.2 &Minangkabau &Arabic &12.6 &12.2 &Swahili &Latin &44.8 &46.7 \\
Moroccan Arabic &Arabic &34.6 &37.4 &Hausa &Latin &38.7 &40.3 &Minangkabau &Latin &33.4 &37.0 &Silesian &Latin &41.2 &44.1 \\
Egyptian Arabic &Arabic &39.0 &41.9 &Hebrew &Hebrew &46.2 &48.5 &Macedonian &Cyrillic &50.6 &53.2 &Tamil &Tamil &35.9 &38.4 \\
Assamese &Bengali &32.5 &35.6 &Hindi &Devanagari &42.1 &45.6 &Plateau Malagasy &Latin &39.7 &41.0 &Tatar &Cyrillic &38.1 &41.0 \\
Asturian &Latin &47.8 &49.9 &Chhattisgarhi &Devanagari &38.0 &41.5 &Maltese &Latin &52.3 &56.2 &Telugu &Telugu &38.3 &42.4 \\
Awadhi &Devanagari &37.9 &41.5 &Croatian &Latin &44.4 &45.2 &Meitei &Bengali &13.8 &13.4 &Tajik &Cyrillic &41.4 &43.0 \\
Central Aymara &Latin &15.1 &14.6 &Hungarian &Latin &44.6 &46.3 &Halh Mongolian &Cyrillic &34.1 &38.4 &Tagalog &Latin &50.9 &52.2 \\
South Azerbaijani &Arabic &27.8 &28.5 &Armenian &Armenian &44.1 &45.5 &Mossi &Latin &15.7 &14.5 &Thai &Thai &40.6 &41.5 \\
North Azerbaijani &Latin &35.1 &37.6 &Igbo &Latin &35.3 &35.9 &Maori &Latin &36.3 &36.9 &Tigrinya &Ge'ez &29.3 &27.5 \\
Bashkir &Cyrillic &34.7 &38.0 &Ilocano &Latin &39.8 &40.5 &Burmese &Myanmar &33.5 &33.6 &Tamasheq &Latin &18.1 &16.8 \\
Bambara &Latin &20.7 &17.8 &Indonesian &Latin &49.7 &51.3 &Dutch &Latin &48.9 &49.6 &Tamasheq &Tifinagh &8.1 &2.0 \\
Balinese &Latin &39.1 &40.2 &Icelandic &Latin &44.2 &45.0 &Norwegian Nynorsk &Latin &53.8 &55.4 &Tok Pisin &Latin &37.7 &37.2 \\
Belarusian &Cyrillic &39.7 &41.6 &Italian &Latin &49.7 &50.8 &Norwegian Bokmål &Latin &53.5 &54.7 &Tswana &Latin &33.8 &36.1 \\
Bemba &Latin &28.0 &29.5 &Javanese &Latin &44.7 &46.1 &Nepali &Devanagari &40.2 &44.2 &Tsonga &Latin &27.9 &28.8 \\
Bengali &Bengali &38.1 &41.7 &Japanese &Japanese &34.5 &36.7 &Northern Sotho &Latin &37.2 &40.5 &Turkmen &Latin &28.2 &31.1 \\
Bhojpuri &Devanagari &34.3 &37.4 &Kabyle &Latin &16.2 &13.7 &Nuer &Latin &12.9 &11.4 &Tumbuka &Latin &32.0 &33.2 \\
Banjar &Arabic &12.3 &10.5 &Jingpho &Latin &15.7 &15.9 &Nyanja &Latin &36.2 &37.8 &Turkish &Latin &40.9 &44.6 \\
Banjar &Latin &36.5 &38.7 &Kamba &Latin &21.5 &21.6 &Occitan &Latin &54.0 &57.4 &Twi &Latin &27.3 &26.8 \\
Standard Tibetan &Tibetan &12.3 &10.9 &Kannada &Kannada &36.8 &40.7 &West Central Oromo &Latin &23.1 &24.7 &Central Atlas Tamazight &Tifinagh &15.9 &4.6 \\
Bosnian &Latin &46.8 &48.4 &Kashmiri &Arabic &24.5 &26.4 &Odia &Oriya &29.0 &40.9 &Uyghur &Arabic &20.9 &19.3 \\
Buginese &Latin &26.7 &27.1 &Kashmiri &Devanagari &20.9 &22.5 &Pangasinan &Latin &34.5 &35.3 &Ukrainian &Cyrillic &48.5 &50.2 \\
Bulgarian &Cyrillic &50.4 &52.7 &Georgian &Georgian &39.5 &41.4 &Eastern Panjabi &Gurmukhi &40.0 &44.6 &Umbundu &Latin &20.8 &19.9 \\
Catalan &Latin &54.5 &56.3 &Central Kanuri &Arabic &9.5 &11.4 &Papiamento &Latin &47.7 &52.4 &Urdu &Arabic &39.3 &41.8 \\
Cebuano &Latin &50.9 &52.2 &Central Kanuri &Latin &18.3 &18.2 &Western Persian &Latin &43.4 &45.8 &Northern Uzbek &Latin &38.6 &41.8 \\
Czech &Latin &49.5 &51.2 &Kazakh &Cyrillic &38.0 &42.6 &Polish &Latin &44.3 &45.0 &Venetian &Latin &45.6 &48.7 \\
Chokwe &Latin &20.4 &19.3 &Kabiyè &Latin &12.7 &11.9 &Portuguese &Arabic &56.5 &58.1 &Vietnamese &Latin &44.3 &44.8 \\
Central Kurdish &Arabic &33.9 &35.7 &Kabuverdianu &Latin &39.2 &45.5 &Dari &Arabic &43.5 &45.3 &Waray &Latin &49.8 &52.3 \\
Crimean Tatar &Latin &32.1 &36.7 &Khmer &Khmer &42.1 &41.1 &Southern Pashto &Arabic &39.0 &40.8 &Wolof &Latin &19.7 &17.3 \\
Welsh &Latin &49.6 &49.1 &Kikuyu &Latin &21.7 &20.5 &Ayacucho Quechua &Latin &21.0 &21.9 &Xhosa &Latin &40.8 &42.3 \\
Danish &Latin &57.4 &58.2 &Kinyarwanda &Latin &37.1 &38.3 &Romanian &Latin &52.2 &54.3 &Eastern Yiddish &Hebrew &51.0 &52.0 \\
German &Latin &55.1 &55.7 &Kyrgyz &Cyrillic &34.1 &37.6 &Rundi &Latin &32.4 &33.5 &Yoruba &Latin &26.3 &28.3 \\
Southwestern Dinka &Latin &17.3 &15.7 &Kimbundu &Latin &22.2 &22.2 &Russian &Cyrillic &46.8 &48.4 &Yue Chinese &Han (Traditional) &36.8 &35.8 \\
Dyula &Latin &17.2 &16.0 &Northern Kurdish &Latin &37.6 &39.7 &Sango &Latin &23.5 &24.5 &Chinese &Han (Simplified) &3.9 &3.9 \\
Dzongkha &Tibetan &3.5 &10.0 &Kikongo &Latin &25.3 &25.3 &Sanskrit &Devanagari &28.6 &29.4 &Chinese &Han (Traditional) &35.3 &35.0 \\
Greek &Greek &46.8 &48.1 &Korean &Hangul &34.9 &36.8 &Santali &Ol Chiki &0.4 &8.4 &Standard Malay &Latin &49.4 &50.7 \\
English &Latin &93.9 &99.6 &Lao &Lao &43.7 &43.8 &Sicilian &Latin &44.9 &49.1 &Zulu &Latin &40.8 &42.1 \\
Esperanto &Latin &52.4 &55.0 &Ligurian &Latin &44.2 &48.7 &Shan &Myanmar &28.1 &15.3 & & & & \\
Estonian &Latin &45.0 &47.1 &Limburgish &Latin &44.0 &48.2 &Sinhala &Sinhala &35.5 &37.8 & & & & \\
Basque &Latin &40.3 &43.5 &Lingala &Latin &27.5 &27.0 &Slovak &Latin &48.7 &50.8 & & & & \\
Ewe &Latin &22.8 &22.2 &Lithuanian &Latin &42.7 &45.1 &Slovenian &Latin &46.4 &47.7 & & & & \\
\bottomrule
\end{tabular}
\caption{chrF++ scores of \texttt{large} mT5 and ByT5 models fine-tuned on 10k German$\rightarrow$English examples. }\label{tab:all_langs_deen}
\end{table}
\end{landscape}

\end{document}